\documentclass[runningheads]{llncs}
\usepackage{graphicx}
\usepackage{subfigure}
\usepackage[colorlinks,
           linkcolor=black,
           anchorcolor=black,
           citecolor=black
           ]{hyperref}
           \usepackage{algorithm}
\usepackage{algorithmic}
\usepackage{multirow}
\usepackage{amsmath}
\usepackage{amssymb}
\usepackage{marvosym}

\begin{document}

\title{Uncertainty-aware Incremental Learning for Multi-organ Segmentation}


\author{Yuhang Zhou 
\and
Xiaoman Zhang 
\and
Shixiang Feng 
\and
Ya Zhang
\and
Yanfeng Wang
}
\authorrunning{Y. Zhou et al.}
\institute{Cooperative Medianet Innovation Center, Shanghai Jiao Tong University 
\email{\{zhouyuhang,xm99sjtu,fengshixiang,ya\_zhang,wangyanfeng\}@sjtu.edu.cn}}

\maketitle
\begin{abstract}
Most existing approaches to train a unified multi-organ segmentation model from several single-organ datasets require simultaneously access multiple datasets during training. In the real scenarios, due to privacy and ethics concerns, the training data of the organs of interest may not be publicly available.
To this end, we investigate a data-free incremental organ segmentation scenario and propose a novel incremental training framework to solve it. We use the pretrained model instead of its own training data for privacy protection.
Specifically, given a pretrained $K$ organ segmentation model and a new single-organ dataset, we train a unified $K+1$ organ segmentation model without accessing any data belonging to the previous training stages. 
Our approach consists of two parts: the background label alignment strategy and the uncertainty-aware guidance strategy. The first part is used for knowledge transfer from the pretained model to the training model. The second part is used to extract the uncertainty information from the pretrained model to guide the whole knowledge transfer process. By combing these two strategies,  more reliable information is extracted from the pretrained model without original training data.
Experiments on multiple publicly available pretrained models and a multi-organ dataset MOBA have demonstrated the effectiveness of our framework.

\keywords{Multi-organ segmentation \and Incremental learning \and  Uncertainty estimation \and Privacy protection.}
\end{abstract}

\section{Introduction}
\begin{figure}[h!]
\centering
\includegraphics[width=0.95\linewidth]{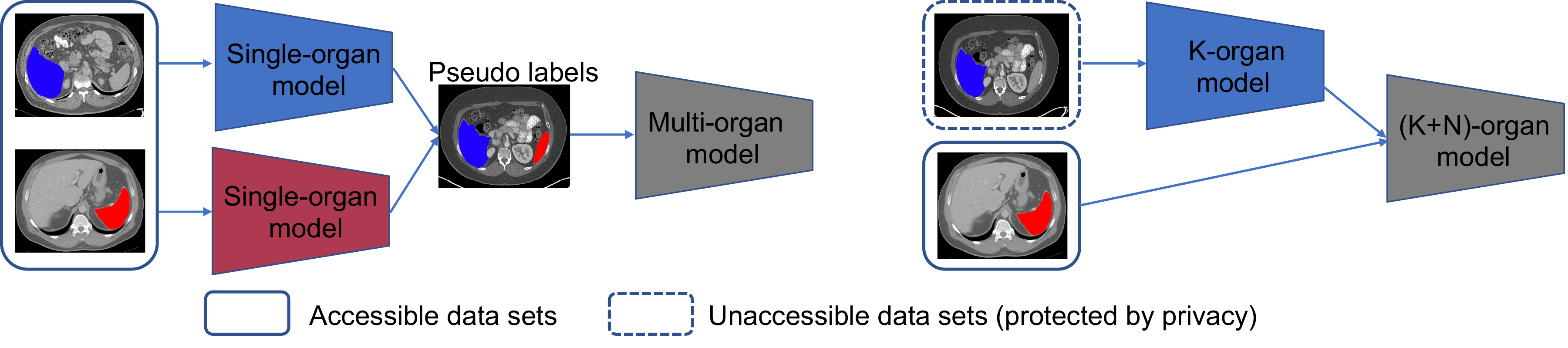}
\caption{The left side of the image represents the pseudo-label based approach, which requires simultaneously access  multiple datasets in the training process. The right side of the image represents our incremental learning setting, which updates the model as a new category is discovered without accessing the data sets from the previous stages.}
\label{com1}
\end{figure}

In recent years, promoted by the large scale publicly available data set, deep convolutional neural networks have made great progress in many application fields, such as image classification\cite{he2016deep}, semantic segmentation\cite{ronneberger2015u} and object detection\cite{ren2015faster}. However, in the field of medical applications, the type and scale of publicly available data are quite limited due to the strict privacy protection.
In this paper, we investigate a  \emph{data-free incremental learning} scenario with application to multi-organ segmentation. Specifically, given a $K$-organ segmentation model and an additional organ segmentation dataset for $N$ new organs, we attempt to learn to segment $K+N$ organs without accessing to the training data of the $K$-organ segmentation model  through \emph{incremental organ segmentation learning}  leveraging the existing models instead of the corresponding data sets.  

The work most related to our study is to train a unified multi-organ segmentation model from several single-organ datasets.
These studies can be roughly divided into two categories: pseudo-label based methods and conditional information based methods. The pseudo-label based methods convert the partial label problem into the full label problem by generating pseudo labels~\cite{huang2020multi,zhou2019prior}.  
The final performance of the model depends heavily on the accuracy of generated pseudo labels. 
Conditional information based methods introduce the conditional control information in the training process to establish a relationship between network parameters and the target organ task~\cite{dmitriev2019learning,zhang2020dodnet}. 
This type of method needs to infer each organ sequentially and is very time-consuming for large number of organs.
Although both types of methods have been proven to be effective, they both require access to multiple datasets during training, which is susceptible to privacy concerns.
This paper proposes an incremental learning framework for multi-organ segmentation.
Different than the existing methods, 
we extend a $K$-organ segmentation model to segment additional organs and require labeled data set only for the new organs (Fig. \ref{com1}).
To the best of our knowledge, we are the first to define this learning setting and introduce incremental learning~\cite{michieli2021knowledge} to solve it.

While incremental learning seems to be a natural solution for this problem, a major challenge in incremental learning is how to learn a new task without forgetting previously learned tasks. Because the training model tends to focus on the new task, which may lead to catastrophic forgetting on old tasks. 
Knowledge distillation\cite{hinton2015distilling}, as an effective transfer learning approach, transfers knowledge from the existing trained teacher model to a new training student model.
However, in our incremental training setting, knowledge distillation cannot be applied directly because the output dimensions of the student model and the teacher model are different. 
Therefore, we propose a background label alignment strategy based on the annotation characteristic of single-organ datasets, \emph{i.e.,} other organs of interest are labeled as the background, which reasonably transforms the probability of the student model output so as to enable knowledge distillation.
Besides, the teacher model can only provide compressed information, which may lead to performance degradation to some extent. We thus use a simple and effective entropy-based evaluation method to mine the uncertainty information from the teacher model to guide the training of the network.
Experiments on multiple publicly available pretrained models and a multi-organ dataset MOBA have shown the effectiveness of our framework.

The main contributions of the paper are summarized as follows: (1) 
We propose a novel training framework for privacy protection, using the trained model instead of the corresponding training data, and successfully apply it to the multi-organ segmentation task.
(2) We design an incremental segmentation framework which combines  background label alignment with uncertainty-aware guidance to transfer knowledge from the teacher model.

\section{Methodology}
Given a trained teacher model $M_K$ to segment $K$ organs, and then given an additional organ segmentation dataset $D_{N}$ which only has the annotation of the $N$ new organs, 
our goal is to train a  student model $M_{K+N}$, which can simultaneously segment $K+N$ organs.
In the rest of the section, we first present how to incrementally extend a trained $K$-organ segmentation model to $K+1$ organs. 
The method can be extended to the setting of multi-organ increment.

Since the data of the $K$-organ segmentat task is not available for incremental organ segmentation, to avoid catastrophic forgetting
of the old tasks, the key is to extract knowledge from the trained model.  This paper employs Knowledge Distillation (KD) to transfer knowledge of the first $K$ organs from the teacher model to the student model in order to maintain  performance on the old task.

\begin{figure}[t]
\includegraphics[width=0.9\linewidth]{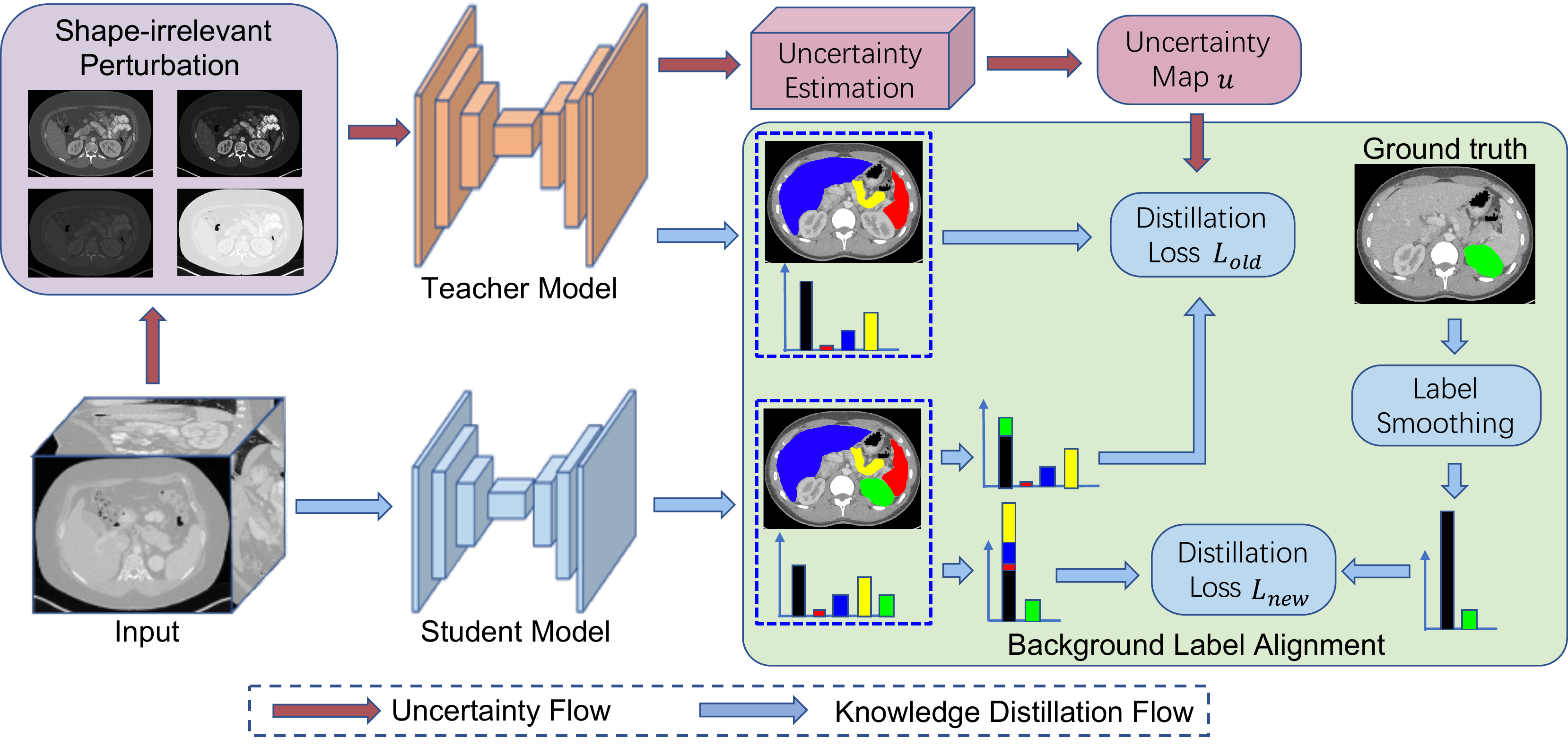}
\centering
\caption{Our proposed framework. The blue arrows represent the process of knowledge distillation where the outputs of the teacher model and student model are matched by our background label alignment strategy. The red arrows represent the process of uncertainty-aware guidance. The input patches after many times shape-invariable disturbances are input to the teacher model for obtaining a more reliable uncertainty map which can guide the training of the whole network.} 
\label{model}
\end{figure}

\subsection{The KD framework for Incremental Organ Segmentation}
The typical knowledge distillation approach is to apply a KD loss at the output level between the teacher model and the student model. Specifically,
the outputs of the teacher network $p^t(k)$ and that of the student network $p^s(k)$ are:
\begin{equation}
\begin{aligned}
& p^t(k) &= &\frac{exp(z_k^t)}{\sum_{i=0}^Kexp(z_i^t)},k\in \{0,...,K\}\\
& p^s(k)&=&\frac{exp(z_k^s)}{\sum_{i=0}^{K+1}exp(z_i^s)},k\in \{0,...,K+1\},
\end{aligned}
\end{equation}
where $z^t$ and $z^s$ are the output logits of the teacher model and the student model, respectively. 
The standard KD loss function can be written as
\begin{equation}
\label{eqn_1}
\begin{aligned}
L_{KD}=(1-\alpha)H(q,p^s)+\alpha D_{KL}(p^t,p^s)
\end{aligned}
\end{equation}
where $q(k)$ is the ground-truth distribution over the labels, $D_{KL}$ is the Kullback-Leibler divergence, and $H(q,p^s)=-\sum^K_{k=0}q(k)log(p^s(k))$ is the cross-entropy loss. 
However, for the incremental organ segmentation, the output dimensions of the teacher model and the student model are different, $K+1$ vs. $K+2$ due to the new incremental class, so the knowledge distillation loss cannot be directly applied. 

\subsection{Background label alignment}
To enable knowledge distillation from the teacher to student with different output dimension, we design a background label alignment strategy, as shown in the Fig.\ref{model}.
Since the $(K+1)th$ organ is regarded as the background class in the output of the teacher model, 
we take $p^t(0)$ as the target to supervise the background and the $(K+1)th$ organ, \emph{i.e.,}  $p^s(0)+p^s(K+1)$.
The purpose of this strategy is to retain the similarity relationship of the first $K$ organs relative to the real background class and the new increment class. The probability transformation and distillation loss on the old tasks, \emph{i.e.,} segmenting the first $K$ organs, can be written as
\begin{equation}
\label{eqn_2}
\begin{aligned}
\hat{p}_{old}^s(k)= \left \{
\begin{array}{ll}
p^s(k),  & for \  k \in \{1,...,K\}\\
1 - \sum_{i=1}^K p^s(i),     & for \  k =  0
\end{array}
\right.
\end{aligned}
\end{equation}

\begin{equation}
\label{eqn_3}
\begin{aligned}
L_{old} = (1-\alpha) H(\arg\max (p^t),\hat p_{old}^s) + \alpha D_{KL}(p^t,\hat p_{old}^s).
\end{aligned}
\end{equation}
where $H(\arg\max(p^t),\hat p_o^s)$ can be used to speed up convergence, especially for small organs like pancreas, which is observed experimentally.

For the annotated new organ dataset, its background class also contains the first $K$ organs, so we also apply the background label alignment strategy to it. 
Denote the ground truth labels for the new organ dataset as $g$
and the probability transformation and distillation loss for the new task can be written as 
\begin{equation}
\label{eqn_4}
\begin{aligned}
\hat{p}_{new}^s(k)= \left \{
\begin{array}{ll}
\sum_{i=0}^{K} p^s(i),  & for \ k = 0\\
p^s(K+k),     & for \ k  = 1
\end{array}
\right.
\end{aligned}
\end{equation}

\begin{equation}
\label{eqn_5}
\begin{aligned}
L_{new} = D_{KL}(\hat p_{new}^s,g).
\end{aligned}
\end{equation}

\subsection{Uncertainty-aware guidance}
Since the prediction of the teacher model on the new organ dataset is easily affected by the domain shift and the generalization of the model itself, the supervision provided to the student model may be noisy, making it difficult for the student to learn from the teacher. 
In order to obtain more reliable knowledge from the teacher model, we use a simple and effective entropy-based evaluation method to estimate uncertainty of the teacher model which is further used to guide the training of the whole network for the old tasks.

As shown in the Fig.\ref{model}, we do $Q$ times shape-irrelevant perturbations 
for each input image $x$. 
The perturbations are one or several transformations randomly selected according to a specified probability from a shape-irrelevant data augmentation pool which contains contrast transformation, brightness transformation, Gaussian blur and Gaussian noise.
By integrating the predictions of the teacher model on the same input with different perturbations, we obtain a more reliable uncertainty map.
Specifically, for each training example $x$ and its augmented set $\{x_1,...,x_Q\}$, the teacher model outputs a set of softmax probability maps $\{y_1,...,y_Q\}$  and then the uncertainty map can be computed with
\begin{equation}
\label{eqn_6}
\begin{aligned}
 u = -\sum_k \bar{y}_k log\bar{y}_k
\end{aligned}
\end{equation}
where $\bar{y}_k = \frac{1}{Q}\sum_q y_q$. We use the result of uncertainty evaluation as the weight of $L_{old}$ to guide the KD training.

\subsection{Overall training loss} The overall training loss of our framework is composed of two parts: the distillation loss of old tasks and the distillation loss of the new task. Then the overall training loss can be written as
\begin{equation}
\label{eqn_all}
\begin{aligned}
L_{total}&=u \lambda_1  H(\arg\max(p^t),\hat p_{old}^s) + u \lambda_2 D_{KL}(p^t,\hat p_{old}^s) + \lambda_3 D_{KL}(\hat p_{new}^s,g_s),
\end{aligned}
\end{equation}
where $\lambda_1$, $\lambda_2$ and $\lambda_3$ denote weights of different loss terms.

\section{Experiments and Results}
In order to show the generalization of our method, we select different pretrained teacher models to construct three scenarios on MOBA\cite{gibson2018automatic} dataset.
Specifically, our three scenarios are ``Spleen+Kidney'', ``Liver+ Pancreas" and ``$\{$Spleen,Liver,\\Pancrea$s\}$+Kidney" respectively. The item before the plus sign represents the pretrained teacher model with the ability to segment the corresponding organs, and the item after the plus sign represents the new incremental organ.

MOBA\cite{gibson2018automatic} is a multi-organ dataset with a total of 90 samples, and each sample is labeled with 8 organs: spleen, left kidney, gallbladder, esophagus, liver, stomach, pancreas and duodenum respectively. We can convert it to corresponding single-organ datasets according to our experimental scenarios and then use them as our new task.
Specifically, under the ``Spleen+Kidney" scenario, our pretrained teacher model\footnotemark[1] is a publicly downloadable model trained on MSD Spleen\cite{simpson2019large} by nnunet\cite{isensee2018nnu}. 
Similarly, under the ``Liver+Pancreas" scenario, we select the model trained by nnunet\cite{isensee2018nnu} on LiTS\cite{bilic2019liver} dataset, where the model can also be publicly downloaded\footnotemark[1]. MSD Spleen\cite{simpson2019large} dataset includes 41 training cases with spleen annotations and LiTS\cite{bilic2019liver} dataset includes 131 training CT cases with liver annotations.
\footnotetext[1]{\href{https://zenodo.org/record/4003545}{https://zenodo.org/record/4003545}}

It is worth noting that there is no directly publicly available pretrained model under ``$\{$Spleen,Liver,Pancreas$\}$+Kidney" scenario. For this reason we use the nnunet framework\cite{isensee2018nnu} to train a teacher model by self-training\cite{papandreou2015weakly} on the mixed dataset $\{$MSD Spleen,LiTS,NIH Pancreas$\}$. NIH Pancreas dataset \cite{roth2016data} consists of 82 abdominal contrast enhanced 3D CT images. All comparative experiments are validated under these three scenarios. We use the Dice-Score-Coefficient(DSC) as our evaluation metric : $DCS(P,G)=\frac{2\times \left |  P\times G\right |}{\left | P \right | + \left | G \right |}$  where $P$ is the binary prediction of the student model and $G$ is the ground truth.

In addition, in the preprocessing stage, we need to resample the training data of MOBA\cite{gibson2018automatic} according to the spacing used by the pretrained teacher model. 

\subsection{Implementation details}
We apply label smoothing to the ground-truth label of the new organ training set where label smoothing can be seen as a kind of regularization so that the student model can be trained better\cite{yuan2020revisiting}. Specifically, for our $K+1$ experiments, we use 0.7 and 0.3 to replace the original hard label for the new single-organ data set. And for the uncertainty-aware guidance, we set $Q = 6$ in our experiments. $\lambda_1$, $\lambda_2$ and $\lambda_3$ are 1, 20, 20, respectively.

We use nnunet\cite{isensee2018nnu} as our basic framework for the 3D incremental segmentation training with a minibatch of 2.
As the optimizer of network training, Adam's initial learning rate is 0.00003 and the weight decay is 0.00003. During training, the parameters of the teacher model are fixed, and only the parameters of the student model are optimized. All datasets are divided into training set and test set with a 4:1 ratio. The implementation of each experiment group is completely consistent except for the method part. 

\subsection{Ablation study}
In this section we evaluate the effectiveness of each component of our method, and the experimental results are shown in Table \ref{table_compare1} and Table \ref{table_compare2}.
We use ``w all datasets'' to represent ``train with all datasets $\{D_1,D_2,...,D_K,D_{K+1}\}$ by self-training\cite{papandreou2015weakly}'' and use ``w full labels'' to represent 
``directly train with $D_{K+1}$  which has annotations of all the $K+1$ organs''.

We notice that in the ``Spleen+Kidney" task and ``Liver+ Pancreas" task, ``w all datasets'' is worse than ``w full labels'' (89.55$\%$ vs. 95.77$\%$ and 86.16$\%$ vs. 87.03$\%$), and in the ``$\{$Spleen,Liver,Pancreas$\}$+Kidney" task, they are comparable (91.39$\%$ vs. 91.40$\%$).  We speculate that this is because the negative impact of noise from generated pseudo labels decreases as the amount of noisy data increases, and it can even hurt the performance when the amount of noisy data is not enough. Under the three scenarios, the number of samples increased by ``w all datasets" is 41, 131 and 254 respectively.

By using the background label alignment strategy and the uncertainty-aware guidance strategy, our method can achieve comparable or slightly better performance than ``w all datasets'' and ``w full labels'' (95.70$\%$ vs.  89.55$\%$ and 95.77$\%$, 87.24$\%$ vs. 86.16$\%$ and 87.03$\%$, 91.73$\%$ vs. 91.39$\%$ and 91.40$\%$). This may be due to the fact that the pretrained teacher model contains additional information learned from previously used datasets. Further, these information can be extracted and leveraged to make up and enhance performance by our method. Our approach not only achieves privacy protection and task increment, but also achieves comparable performance on almost all tasks.

When only the background label alignment strategy is used (denotes as ``w/o uncertainty''), the performance suffers from varying degrees of degradation on old tasks (95.46$\%$ vs. 92.28$\%$, 96.06$\%$ vs. 94.75$\%$, 95.91$\%$ vs. 95.90$\%$, 79.08$\%$ vs. 78.93$\%$). This indicates that our uncertainty-aware guidance strategy can provide more reliable uncertainty map for knowledge transfer. We notice it also provides the performance gain for the new task (95.93$\%$ vs. 93.03$\%$, 78.42$\%$ vs. 77.34$\%$, 95.87$\%$ vs. 95.11$\%$). We infer that the uncertainty estimation can also be regarded as a regularization to make the network converge to a better optimal solution.

\subsection{Comparion with state-of-the-art}
We also compare our method with the state-of-the-art approach MiB\cite{cermelli2020modeling} which is an effective incremental framework for semantic segmentation and the experimental results are shown in Table \ref{table_compare1} and Table \ref{table_compare2}. 
Since MiB\cite{cermelli2020modeling} is designed for 2D natural scene, we adjust it to adapt our 3D task.
Then we can see that our method has better performance than theirs (95.70$\%$ vs. 91.32$\%$, 87.24$\%$ vs. 85.82$\%$, 91.73$\%$ vs. 90.72$\%$). This may be because we take into account another important topic in the medical field i.e. uncertainty and combine it with incremental learning to form a unified framework.

\begin{figure}[h!]
\centering
\includegraphics[width=0.95\linewidth]{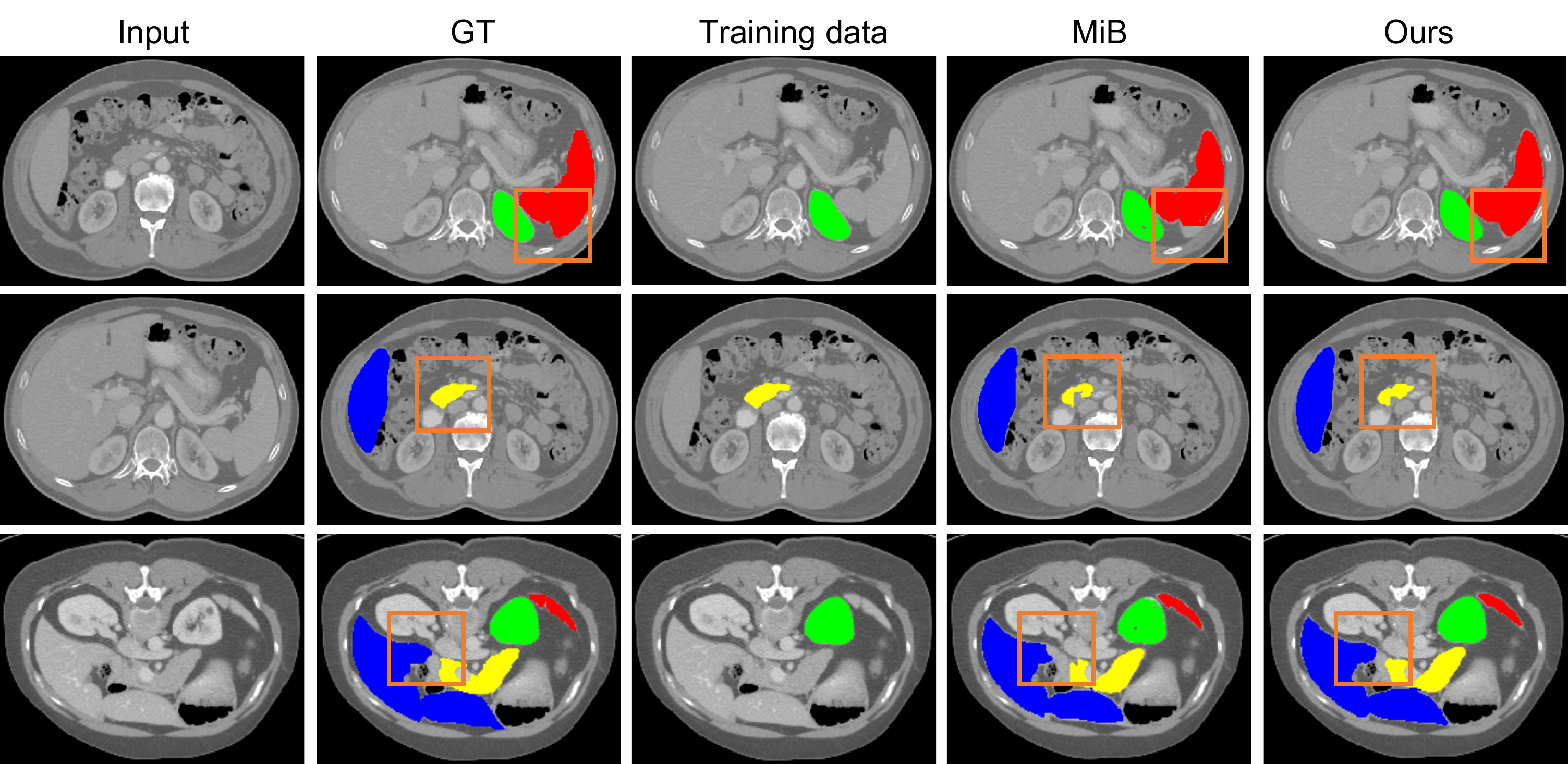}
\caption{Visual results under three different scenarios. Top to bottom: ``Spleen+Kidney" scenario, ``Liver+ Pancreas" scenario and ``$\{$Spleen,Liver,Pancreas$\}$+Kidney" scenario.}
\label{result}
\end{figure}

Experiments under three different scenarios demonstrate the effectiveness and robustness of our framework. Some qualitative results are shown in Figure \ref{result}. We can see that our method shows better results  than MiB\cite{cermelli2020modeling}.

\begin{table}[h]\scriptsize
\centering
\caption{DSC($\%$) comparison on ``$\{$Spleen,Liver,Pancreas$\}$+Kidney" task.}
\setlength{\tabcolsep}{8pt}
\begin{tabular}{c|cccc|c}
\hline
& \multicolumn{5}{c}{$\{$Spleen+Liver+Pancreas$\}$ + kidney} \\
\hline
& Spleen & Liver & Pancreas & Kidney & Average \\
\hline
w/o uncertainty & 95.90 & \textbf{96.12} & 78.93 & 95.11 & 91.52\\
w all datasets  & 95.54 & 95.00 & \textbf{79.72} & 95.31 & 91.39\\
w full labels  & \textbf{96.15} & 95.73 & 78.32 & 95.39 & 91.40\\
\hline
MiB\cite{cermelli2020modeling} & 95.26  & 95.26 & 77.39  &  94.97 & 90.72 \\
\hline
\textbf{ours}& 95.91 & 96.07& 79.08 & \textbf{95.87} & \textbf{91.73}\\
\hline
\end{tabular}
\label{table_compare1}
\end{table}

\begin{table}[h]\scriptsize
\centering
\caption{DSC($\%$) comparison on ``Spleen+Kidney" task and ``Liver+ Pancreas" task.}
\setlength{\tabcolsep}{8pt}
\begin{tabular}{c|cc|c|cc|c}
\hline
& \multicolumn{3}{c|}{Spleen + kidney} &  \multicolumn{3}{c}{Liver + pancreas} \\
\hline
& Spleen & Kidney & Average & Liver & Pancreas & Average \\
\hline
w/o uncertainty & 92.28 & 93.03 & 92.66 & 94.75 & 77.34 & 86.05\\
w all datasets  & 87.16 & 91.93 & 89.55 & 95.95 & 76.36 & 86.16\\
w full labels  & \textbf{96.15} & 95.39 & \textbf{95.77} & 95.73 & 78.32 & 87.03\\
\hline
MiB\cite{cermelli2020modeling} &  89.72 & 92.92 &  91.32 & 95.18  & 76.46  & 85.82 \\
\hline
\textbf{ours}& 95.46 & \textbf{95.93} & 95.70 & \textbf{96.06} & \textbf{78.42} & \textbf{87.24}\\
\hline
\end{tabular}
\label{table_compare2}
\end{table}

\section{Conclusion}
We propose a new approach for privacy protection, using open source models instead of open source data. 
Specifically, for the abdominal multi-organ segmentation task, we design a novel incremental segmentation framework based on a background label alignment strategy and an uncertainty-aware guidance strategy for training a unified  multi-organ segmentation model. The framework can be easily extended to any number of organ segmentation tasks and doesn't require any data from previous training stages.
Our framework performs well in a variety of experimental scenarios, demonstrating the potential of our approach for multi-organ segmentation tasks.

\bibliographystyle{splncs04}
\bibliography{paper848}
\end{document}